\documentclass[10pt,twocolumn,letterpaper]{article}

\usepackage{iccv}              % To produce the CAMERA-READY version
\definecolor{iccvblue}{rgb}{0.21,0.49,0.74}
\usepackage[pagebackref,breaklinks,colorlinks,allcolors=iccvblue]{hyperref}
\usepackage{multirow}
\usepackage{makecell}
\usepackage{CJKutf8}
\def\paperID{5815} % *** Enter the Paper ID here
\def\confName{ICCV}
\def\confYear{2025}

\title{PriOr-Flow: Enhancing \textcolor{red}{Pri}mitive Panoramic Optical Flow \\with \textcolor{red}{Or}thogonal View}

\author{Longliang Liu,
~~Miaojie Feng,
~~Junda Cheng,
~~Jijun Xiang,
~~Xuan Zhu,
~~Xin Yang\footnotemark[2]\\
[2mm]
School of EIC, Huazhong University of Science and Technology\\
\tt\small \{longliangl, fmj, jundacheng, jijunx, xuanzhu, xinyang2014\}@hust.edu.cn
}

\begin{document}
\begin{CJK}{UTF8}{gbsn}
\maketitle
\begin{abstract}
\vspace{-20pt}
\renewcommand{\thefootnote}{\fnsymbol{footnote}}
\footnotetext[2]{Corresponding author.}

Panoramic optical flow enables a comprehensive understanding of temporal dynamics across wide fields of view. However, severe distortions caused by sphere-to-plane projections, such as the equirectangular projection (ERP), significantly degrade the performance of conventional perspective-based optical flow methods, especially in polar regions. To address this challenge, we propose PriOr-Flow, a novel dual-branch framework that leverages the low-distortion nature of the orthogonal view to enhance optical flow estimation in these regions. Specifically, we introduce the Dual-Cost Collaborative Lookup (DCCL) operator, which jointly retrieves correlation information from both the primitive and orthogonal cost volumes, effectively mitigating distortion noise during cost volume construction. Furthermore, our Ortho-Driven Distortion Compensation (ODDC) module iteratively refines motion features from both branches, further suppressing polar distortions. Extensive experiments demonstrate that PriOr-Flow is compatible with various perspective-based iterative optical flow methods and consistently achieves state-of-the-art performance on publicly available panoramic optical flow datasets, setting a new benchmark for wide-field motion estimation. The code is publicly available at: \textcolor{magenta}{https://github.com/longliangLiu/PriOr-Flow}.

\end{abstract}    
\vspace{-15pt}
\section{Introduction}
\label{sec:intro}

Optical flow estimates a dense motion vector field between consecutive video frames, capturing the apparent displacement of pixels on the image plane. As a fundamental problem in computer vision, it plays a pivotal role in dynamic scene understanding and supports a wide range of applications, including frame interpolation \cite{frame_interpolation0, frame_interpolation1, frame_interpolation2}, video inpainting \cite{video_inpainting0, video_inpainting1, video_inpainting2}, autonomous driving \cite{kitti12, auto_dirving, sdv-loam, sr-lio, sr-livo}, and 3D reconstruction and synthesis \cite{reconstruction0, reconstruction1, omnisplat, voxel-svio}.

%%%%%%%%%%%%%%%%%%%%%%%%%%%%%%%%%%%%%%%%%%%%%%%%%%%%%%%%%%%
%% Sota比较
\begin{figure}[t]
    \centering
    \includegraphics[width=\columnwidth]{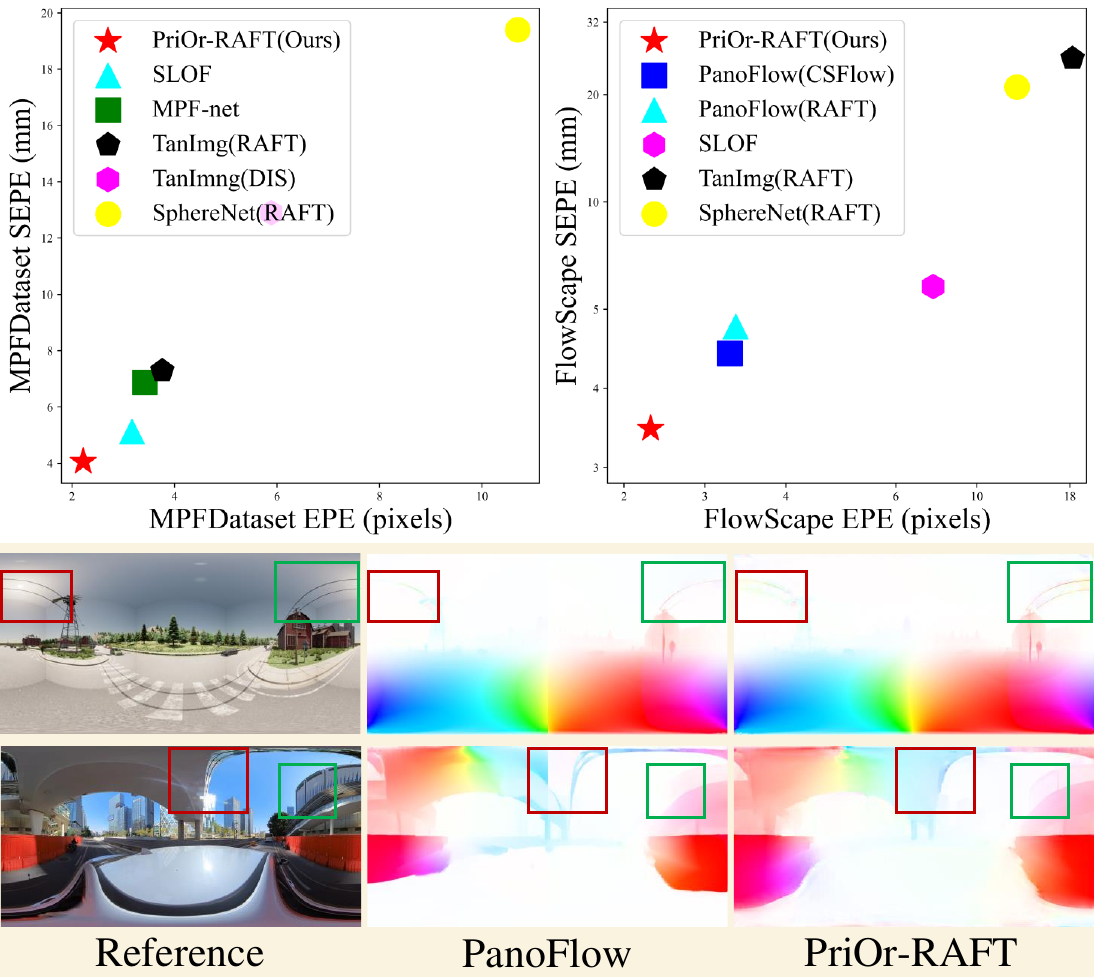}
    \vspace{-15pt}
    \caption{\textbf{Row 1:} Comparisons with state-of-the-art panoramic optical flow estimation methods on MPFDataset~\cite{MPF-net} and FlowScape~\cite{panoflow}. \textbf{Row 2:} Visual comparison on the virtual dataset FlowScape~\cite{panoflow}. \textbf{Row 3:} Visual comparison on the real dataset ODVista~\cite{odvista}. Our method performs well in the polar regions.}
    \label{fig:compare_sota}
    \vspace{-20pt}
\end{figure}
%%%%%%%%%%%%%%%%%%%%%%%%%%%%%%%%%%%%%%%%%%%%%%%%%%%%%%%%%%

In the realm of perspective imagery, recent approaches \cite{raft, GMA, GMFlowNet, skflow, cheng2022region, gmflow, flowformer, coatrsnet} have achieved remarkable success on well-established benchmarks. However, the increasing availability of panoramic cameras has created a pressing demand for panoramic optical flow estimation, which enables comprehensive motion understanding across wide fields of view. Among various projection formats, equirectangular projection (ERP) remains the most prevalent. Yet, by mapping a spherical image onto a 2D plane, ERP introduces severe distortions, particularly near the poles. As a result, applying conventional optical flow methods designed for perspective images directly to panoramic images leads to substantial performance degradation.

To address this challenge, several works have explored specialized architectures for panoramic optical flow, which can be broadly categorized into three groups: weight transformation-based, tangent plane-based, and ERP-based methods.\textit{ Weight transformation-based methods}~\cite{LiteFlowNet360, omniflownet} adapt perspective-trained optical flow models to ERP images via weight transformations, eliminating the need for extensive panoramic optical flow datasets. However, these additional transformation layers introduce efficiency and portability concerns~\cite{slof}. \textit{Tangent plane-based methods}~\cite{TanImg} mitigate distortions by projecting spherical images onto multiple tangent planes, allowing perspective optical flow models to operate in a locally undistorted manner. Nevertheless, this approach suffers from inter-plane discontinuities and cross-plane motion inconsistencies. \textit{ERP-based methods} directly process ERP images and employ techniques such as deformable convolution~\cite{dcn} and representation learning~\cite{siamese} to compensate for projection distortions. However, ERP-induced distortions are most severe in polar regions, where existing approaches exhibit frequent and significant errors, as shown in Fig.~\ref{fig:compare_sota}. Despite these advancements, no existing method explicitly addresses the distortion effects in polar regions, leaving a critical gap in panoramic optical flow estimation.

%%%%%%%%%%%%%%%%%%%%%%%%%%%%%%%%%%%%%%%%%%%%%%%%%%%%%%%%%%%
%% 球面旋转示意图
\begin{figure}[t]
    \centering
    \includegraphics[width=\columnwidth]{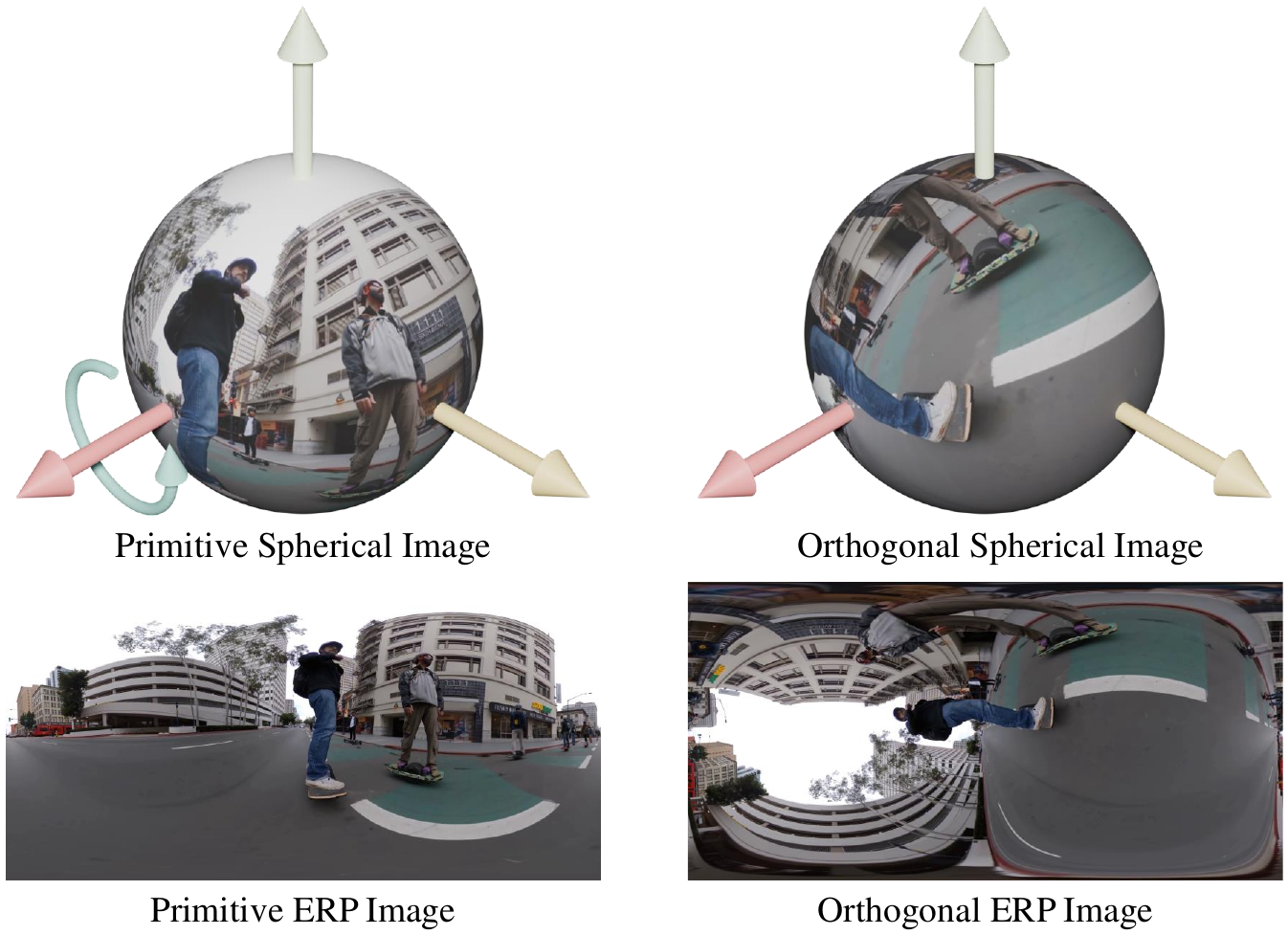}
    \vspace{-20pt}
    \caption{The orthogonal view is obtained by rotating the spherical image by $90^{\circ}$ around the x-axis (or y-axis). The first row illustrates the spherical rotation process, while the second row shows the corresponding ERP projection images.}
    \label{fig:sphrotate}
    \vspace{-20pt}
\end{figure}
%%%%%%%%%%%%%%%%%%%%%%%%%%%%%%%%%%%%%%%%%%%%%%%%%%%%%%%%%%

% In contrast to other ERP-based methods, in this paper, we propose PriOr-Flow, a novel dual-branch~\cite{dual_branch_dehazing, dual_branch_super, elfnet, monster} panoramic optical flow estimation approach that leverages the low-distortion prior of the orthogonal view to compensate for optical flow recovery in the polar regions. As mentioned in \cite{osrt, sph_weighted}, due to the non-uniform sampling of the ERP projection, the extent of distortion varies with latitude following a cosine pattern
% % latitude $\phi$ following a cosine pattern:
% % \begin{equation}
% %     \mathcal{D} = \frac{\lvert dx dy \rvert}{\cos\phi\lvert d\theta d\phi\rvert} = \frac{1}{\cos\phi}
% %     \label{equ:dist}
% % \end{equation}
% % where all symbols follow the definitions in Eq.~\eqref{equ:ERP}, 
% and the distortion reaches its maximum at the polar regions, as shown in Fig.~\ref{fig:dist}(a). However, as illustrated in Fig.~\ref{fig:dist}, the orthographic view, obtained by rotating the panoramic image by $90^{\circ}$ in the spherical space (as shown in Fig.~\ref{fig:sphrotate}), exhibits a distortion distribution opposite to that of the primitive view, featuring minimal distortion in the polar regions. Throughout the iterative process, the primitive branch of PriOr-Flow continuously leverages the low-distortion prior of the polar regions from the orthogonal branch, significantly enhancing the accuracy of optical flow prediction in these areas.

\begin{figure}[t]
    \centering
    \includegraphics[width=\columnwidth]{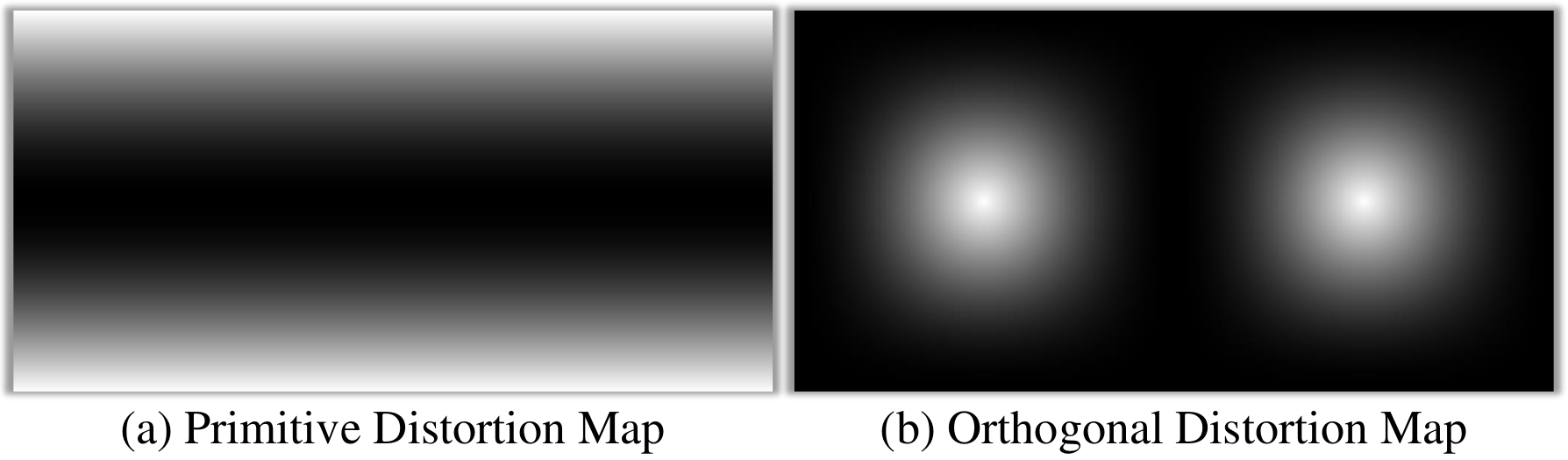}
    \vspace{-20pt}
    \caption{Distortion maps of the primitive view and the orthogonal view, where brighter regions indicate higher distortion levels.}
    \label{fig:dist}
    \vspace{-20pt}
\end{figure}

To address the limitations of existing ERP-based methods, we introduce PriOr-Flow, a novel dual-branch panoramic optical flow estimation framework that uniquely exploits the low-distortion prior of the orthogonal view to explicitly compensate for severe distortions in polar regions~\cite{dual_branch_dehazing, dual_branch_super, elfnet, monster, afnet}. As noted in \cite{osrt, sph_weighted}, the non-uniform sampling of ERP projection results in latitude-dependent distortions following a cosine pattern, with the most severe distortions occurring at the poles (Fig.~\ref{fig:dist}(a)). This distortion significantly degrades optical flow accuracy in these regions. To mitigate this issue, we leverage the orthogonal view, obtained by rotating the panoramic image by $90^{\circ}$ in spherical space (Fig.\ref{fig:sphrotate}). Unlike the primitive ERP view, the orthogonal view exhibits an opposite distortion pattern, featuring minimal distortions in the polar regions (Fig.\ref{fig:dist}). During iterative refinement, PriOr-Flow’s primitive branch adaptively incorporates the low-distortion prior from the orthogonal branch, significantly enhancing optical flow estimation in polar regions.

% Specifically, to mitigate the distortion noise introduced during the cost volume construction, we propose the Dual-Cost Collaborative Lookup (DCCL) operator. This operator utilizes the currently estimated optical flow to jointly retrieve correlation information from both the primitive cost volume and the orthogonal cost volume. By doing so, it avoids the scenario where a single cost volume retrieval contains excessive noise, thereby preventing erroneous optical flow recovery. To further exploit the low-distortion prior of the orthogonal view, we introduce the Ortho-Driven Distortion Compensation (ODDC) module. During the iterative process, this module explicitly guides the fusion of motion features from both the primitive and orthogonal branches by leveraging the confidence information of the primitive and orthogonal optical flows. By doing so, the more accurate motion features of the orthogonal branch in the polar regions are integrated into the original branch, enabling the primitive branch to achieve truly ``omnidirectional'' optical flow estimation. Additionally, we demonstrate the module's effectiveness and broad applicability by integrating it into various iterative network architectures. The entire framework is unified under the name PriOr-Flow.

Specifically, to mitigate distortion noise introduced during cost volume construction, we propose the Dual-Cost Collaborative Lookup (DCCL) operator. This operator leverages the estimated optical flow to jointly extract correlation information from both the primitive and orthogonal cost volumes. By incorporating both sources, DCCL prevents excessive noise in individual cost volume retrievals, thereby reducing optical flow estimation errors. To further harness the low-distortion prior of the orthogonal view, we introduce the Ortho-Driven Distortion Compensation (ODDC) module. This module explicitly guides the fusion of motion features across both branches based on their confidence estimates, ensuring that high-quality motion features from the orthogonal branch—especially in polar regions—are effectively integrated into the primitive branch. As a result, the primitive branch benefits from enhanced accuracy, ultimately enabling robust omnidirectional optical flow estimation. Additionally, we demonstrate the module's effectiveness and broad applicability by integrating it into various iterative network architectures. Together, these components form a cohesive framework, PriOr-Flow, which establishes a new benchmark for panoramic optical flow estimation.

%%%%%%%%%%%%%%%%%%%%%%%%%%%%%%%%%%%%%%%%%%%%%%%%%%%%%%%%%%%
%% Distortion Map
%%%%%%%%%%%%%%%%%%%%%%%%%%%%%%%%%%%%%%%%%%%%%%%%%%%%%%%%%%

% We evaluate our PriOr-Flow on the publicly available panoramic optical flow datasets, the MPFDataset~\cite{MPF-net} and FlowScape~\cite{panoflow}. As shown in Fig.~\ref{fig:compare_sota}, our PriOr-RAFT achieved state-of-the-art performance. Specifically, PriOr-RAFT achieves improvements of 30.0\% and 29.6\% in the EPE metric on the two datasets, respectively. Moreover, Fig.~\ref{fig:compare_sota} demonstrates that our method performs better in the polar regions. We also compare our method with the previous best approach in real-world scenarios, demonstrating that our method generalizes well to practical environments. Extensive ablation experiments further validate the effectiveness of each component we proposed.  

% In summary, our contributions are as follows:
% \begin{itemize}
%     \item We introduce PriOr-Flow, a novel dual-branch panoramic optical flow estimation method that leverages the low-distortion prior of the orthogonal view to compensate for optical flow recovery in the polar regions.
%     \item The proposed DCCL and ODDC modules effectively harness low-distortion priors of the orthogonal view.
%     \item We verify the universality of our method on several iterative optical flow methods.
%     \item Our method achieves state-of-the-art performance across multiple datasets and demonstrates strong generalization capabilities in real-world scenarios.
% \end{itemize}

We evaluate PriOr-Flow on publicly available panoramic optical flow datasets, MPFDataset~\cite{MPF-net} and FlowScape~\cite{panoflow}. As shown in Fig.~\ref{fig:compare_sota}, PriOr-RAFT establishes a new state-of-the-art, significantly outperforming previous methods. Specifically, PriOr-RAFT reduces the EPE metric by 30.0\% and 29.6\% on the two datasets, respectively, demonstrating a substantial performance gain. Moreover, Fig.~\ref{fig:compare_sota} highlights that our method significantly improves accuracy in the polar regions. We further compare PriOr-Flow with the previous best approach in real-world scenarios, showing that it generalizes well across diverse practical environments. Extensive ablation studies validate the effectiveness of each proposed component.  

\noindent In summary, our key contributions are:
\begin{itemize}
    \item We propose PriOr-Flow, a novel dual-branch framework that exploits the orthogonal view's low-distortion prior to enhance optical flow estimation in polar regions.
    \item We introduce the DCCL and ODDC modules to explicitly integrate low-distortion priors from the orthogonal view, reducing estimation errors in high-distortion areas.
    \item We demonstrate the broad applicability of PriOr-Flow by integrating it with multiple iterative optical flow architectures.
    \item PriOr-Flow sets a new state-of-the-art on multiple panoramic optical flow benchmarks and exhibits strong generalization to real-world applications.
\end{itemize}

\vspace{-5pt}
\section{Related Work}
\label{sec:related works}

% \subsection{Perspective Optical Flow Estimation}
% Recent advances \cite{flownet, pwc-net, raft, GMA, GMFlowNet, gmflow, flowformer} in optical flow estimation have been driven by deep learning architectures. FlowNet \cite{flownet} pioneered the use of convolutional neural networks (CNNs) to treat optical flow as a regression task, directly predicting dense motion fields from image pairs. Building on this, PWC-Net \cite{pwc-net} introduced a feature pyramid and coarse-to-fine strategy, enabling efficient estimation of large displacements by refining predictions progressively from low to high resolutions. Further pushing the boundaries, RAFT \cite{raft} proposed a recurrent framework that constructs a 4D all-pairs cost volume and iteratively updates flow fields at fixed resolution, achieving state-of-the-art accuracy through multi-step refinement. These methods have demonstrated remarkable success in perspective imagery, yet their reliance on planar projection assumptions limits their applicability to omnidirectional data.

\subsection{Perspective Optical Flow Estimation}
\vspace{-2pt}
Recent advances in optical flow estimation have been propelled by deep learning architectures~\cite{flownet, pwc-net, raft, GMA, GMFlowNet, gmflow, skflow, flowformer}. FlowNet~\cite{flownet} pioneered the use of convolutional neural networks (CNNs) to treat optical flow as a regression task, directly predicting dense motion fields from image pairs. Building on this foundation, PWC-Net~\cite{pwc-net} introduced a feature pyramid and coarse-to-fine strategy, enabling efficient estimation of large displacements by refining predictions progressively from low to high resolutions. RAFT~\cite{raft} further advanced the field with a recurrent framework that constructs a 4D all-pairs cost volume and iteratively updates flow fields at fixed resolution, achieving state-of-the-art accuracy through multi-step refinement. These methods have demonstrated significant success in perspective imagery; however, their reliance on planar projection assumptions limits their direct applicability to omnidirectional data.

\vspace{-2pt}

\subsection{Panoramic Optical Flow Estimation}
\vspace{-2pt}
The increasing use of affordable panoramic cameras has created a growing demand for panoramic optical flow estimation, which enables comprehensive temporal understanding across wide fields of view. However, the spherical-to-planar projection in omnidirectional imagery (e.g., equirectangular format) introduces severe geometric distortions~\cite{omniflownet, distortion}, conflicting with the flat-domain priors of conventional optical flow methods. To address these challenges, recent works have developed optical flow algorithms tailored for panoramic images. LiteFlowNet360~\cite{LiteFlowNet360} was the first deep learning-based approach for dense optical flow estimation in panoramic videos, mitigating spherical distortions using a kernel transformer network (KTN~\cite{KTN}) and self-supervised learning. Similarly, OmniFlowNet~\cite{omniflownet} adapts perspective-based CNNs for omnidirectional images without requiring additional training, using distortion-aware convolutions~\cite{equiconv} to align with equirectangular projection. A key limitation of these methods is their reliance on adapting convolutional layers, which introduces computational overhead and reduces flexibility and ease of deployment. TanImg~\cite{TanImg} estimates panoramic optical flow by dividing the spherical image into multiple tangent planes, helping to reduce distortions from equirectangular projection. However, it struggles with large displacements and boundary continuity, limiting its effectiveness. SLOF~\cite{slof} employs a siamese representation learning framework~\cite{siamese} with rotational augmentations and tailored losses to adapt existing flow networks for panoramic optical flow. MPF-Net~\cite{MPF-net} fuses predictions from multiple projection methods, including tri-cylinder and cube padding projections, to complement equirectangular projection. PanoFlow~\cite{panoflow} further enhances panoramic optical flow estimation by incorporating flow distortion augmentation and deformable convolution~\cite{dcn}, as well as a cyclic flow estimation method to handle boundary continuity. While these methods have attempted to reduce the impact of projection distortions from different angles, none have specifically addressed the most severe distortions in the polar regions. Our PriOr-Flow leverages the low-distortion prior from the orthogonal view to compensate for optical flow estimation in these polar regions, significantly improving performance in these critical areas.

\vspace{-2pt}
\subsection{Multi-view Panoramic Approaches}
\vspace{-2pt}
In the field of panoramic image processing, some studies have explored the use of multiple views of panoramic images. \cite{gradient} decomposes panoramic images and stitches results from each part to avoid polar singularities. \cite{E-CNN} simply concatenates optical flows from different views to achieve uniform flow patterns. SLOF~\cite{slof} introduces additional constraints on cross-view optical flow similarity during training to learn view-invariant representations. In contrast, our proposed PriOr-Flow leverages the orthogonal view to address severe distortions in the polar regions of panoramic images.
\vspace{-8pt}

%%%%%%%%%%%%%%%%%%%%%%%%%%%%%%%%%%%%%%%%%%%%%%%%%%%%%%%%%%%%%%%%%%%%%
%%% 网络框图
\begin{figure*}[t]
    \centering
    \includegraphics[width=0.95\linewidth]{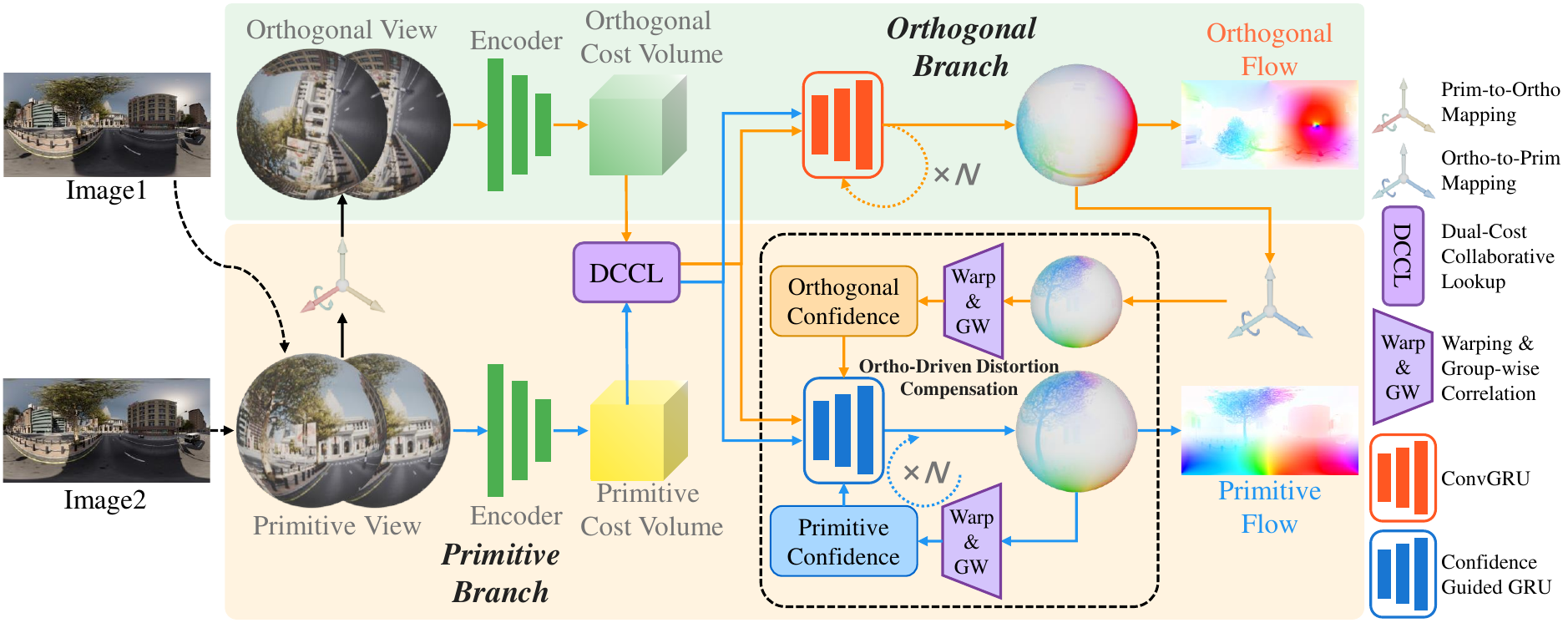}
    \vspace{-5pt}
    \caption{\textbf{Overview of Our proposed PriOr-Flow (PriOr-RAFT version).} The Dual-Cost Collaborative Lookup (DCCL) operator uses the current optical flow to jointly retrieve information from both the primitive and orthogonal cost volumes. In each iteration, the Ortho-Driven Distortion Compensation (ODDC) module aggregates motion features from the orthogonal branch to compensate for the polar regions of the primitive flow.}
    \label{fig:network}
    \vspace{-18pt}
\end{figure*}
%%%%%%%%%%%%%%%%%%%%%%%%%%%%%%%%%%%%%%%%%%%%%%%%%%%%%%%%%%%%%%%%%%%%%

% \subsection{Dual Branch Architecture}
% Currently, many works adopt a dual-branch network architecture, leveraging the complementary characteristics between branches to achieve excellent performance. \cite{dual_branch_dehazing} leverages dual guidance in both the frequency and spatial domains to achieve high-quality dehazing performance. \cite{dual_branch_super} propose a dual prior modulation network, which enhances super-resolution image quality by integrating global structural and semantic information. ELFNet \cite{elfnet} includes a cost-volume-based matching branch and a Transformer-based matching branch, effectively integrating local and global information to achieve superior performance. \cite{monster} proposes a mutual refinement module and effectively fuses matching-driven binocular disparity with semantic-driven relative depth, achieving impressive generalization ability. Unlike other dual-branch architectures, PriOrFlow is the first method to leverage the complementary distortion patterns of orthogonal views and perform iterative flow compensation for mutual refinement. 
\vspace{-8pt}
\section{Method}
\label{sec:method}

% In this section, we present the overall architecture of PriOr-Flow. Since our method can be integrated into various networks, we use PriOr-RAFT (Fig.~\ref{fig:network}) as an example to highlight its key components. We first provide a detailed description of the orthogonal view generation process (Sec.~\ref{sec:orthoview}). Next, we elaborate on the Dual-Cost Collaborative Lookup (DCCL) operator (Sec.~\ref{sec:DCCL}). We then introduce the orthogonal branch (Sec.~\ref{sec:orthobranch}) and the primitive branch (Sec.~\ref{sec:primbranch}); within the latter, we further discuss the Ortho-Driven Distortion Compensation (ODDC) module. Finally, we describe the overall loss function in Sec.~\ref{sec:loss}.

In this section, we present the overall architecture of PriOr-Flow. Since our method can be integrated into various networks, we use PriOr-RAFT (Fig.~\ref{fig:network}) as an example to highlight its key components. We begin by describing the orthogonal view generation process in Section~\ref{sec:orthoview}. Next, we introduce the Dual-Cost Collaborative Lookup (DCCL) operator in Section~\ref{sec:DCCL}. We then detail the primitive and orthogonal branches in Sections~\ref{sec:primbranch} and \ref{sec:orthobranch}, respectively, with a focus on the Ortho-Driven Distortion Compensation (ODDC) module within the primitive branch. Finally, we discuss the overall loss function in Section~\ref{sec:loss}.

\subsection{Orthogonal View Generation}
\label{sec:orthoview}
The equirectangular projection (ERP), the most common format for panoramic imaging, maps spherical coordinates \((\theta, \phi)\) to a 2D plane via a linear transformation \cite{erp_projection, survey}: 

%%%%%%%%%%%%%%%%%%%%%%%%%%%%%%%%%%%%%%%%%%%%%%%%%%%%%%%%%%%%%%%%%%%%%%%%%%
%% ERP投影公式
\begin{equation}
\begin{bmatrix}
u \\
v    
\end{bmatrix} \; = 
\begingroup
\renewcommand{\arraystretch}{1.5}
\begin{bmatrix}
\tfrac{W}{2\pi} & 0 & \frac{W}{2} \\
0 & -\frac{H}{\pi} & \frac{H}{2}
\end{bmatrix}
\endgroup
\begin{bmatrix}
    \theta \\
    \phi \\
    1
\end{bmatrix}
\label{equ:ERP}
\end{equation}
%%%%%%%%%%%%%%%%%%%%%%%%%%%%%%%%%%%%%%%%%%%%%%%%%%%%%%%%%%%%%%%%%%%%%%%%%%

As illustrated in Fig.~\ref{fig:sphrotate}, we generate the orthogonal view by applying spherical rotation to the primitive ERP image. This spherical rotation operation $\mathcal{R}$ is mathematically expressed as:

%%%%%%%%%%%%%%%%%%%%%%%%%%%%%%%%%%%%%%%%%%%%%%%%%%%%%%%%%%%%%%%%%%%%%%%
%% 球面旋转公式
\begin{equation}
    \boldsymbol{\mathrm{x^{\prime}}} = \;
    \mathcal{R}(\theta, \boldsymbol{\mathrm{x}}) = \;
    P^{-1}(R_x(\theta) \cdot P(\boldsymbol{\mathrm{x}}))
\end{equation}
%%%%%%%%%%%%%%%%%%%%%%%%%%%%%%%%%%%%%%%%%%%%%%%%%%%%%%%%%%%%%%%%%%%%%%%

Here, $\boldsymbol{\mathrm{x}}$ represents the coordinates of an ERP pixel, and $\boldsymbol{\mathrm{x^{\prime}}}$ denotes the coordinates after rotation. $P$ is the projection from ERP image coordinates to Cartesian coordinates, and $R_x(\theta)$ is the rotation matrix about the x-axis by angle $\theta$. The orthogonal image is obtained by re-projecting the rotated spherical surface onto the ERP plane:
\begin{equation}
    {I^o} = \;
    T_p^o({I^p}) = \; 
    \left\{\mathrm{Bi}({I^p}, \mathcal{R}(-90^{\circ}, \boldsymbol{\mathrm{x}})) \, 
    | \, \boldsymbol{\mathrm{x}} \in {I^p} \right\}
\end{equation}

Where $I^p$ denotes the primitive image, $I^o$ denotes the orthogonal image, and $\mathrm{Bi}$ represents bilinear interpolation. $T_p^o$ represents the transformation from the primitive view to the orthogonal view, while $T_o^p$ is the inverse transformation.

\subsection{Dual-Cost Collaborative Lookup (DCCL)}
\label{sec:DCCL}
Due to projection distortions, significant noise may be encoded in the cost volume, especially in the polar regions, and the retrieved noisy correlation cues lead to erroneous optical flow field recovery. To address this issue, we propose the \emph{dual-cost collaborative lookup} (DCCL) operator, which performs joint retrieval from both the primitive correlation pyramid $\{C^p_i\}$ and the orthogonal correlation pyramid $\{C^o_i\}$ (defined later in Section \ref{sec:orthobranch}) to mitigate the impact of noise in the cost volume.

%%%%%%%%%%%%%%%%%%%%%%%%%%%%%%%%%%%%%%%%%%%%%%%%%%%%%%%%%%%
%% 双边查找
\begin{figure}[t]
    \centering
    \includegraphics[width=\columnwidth]{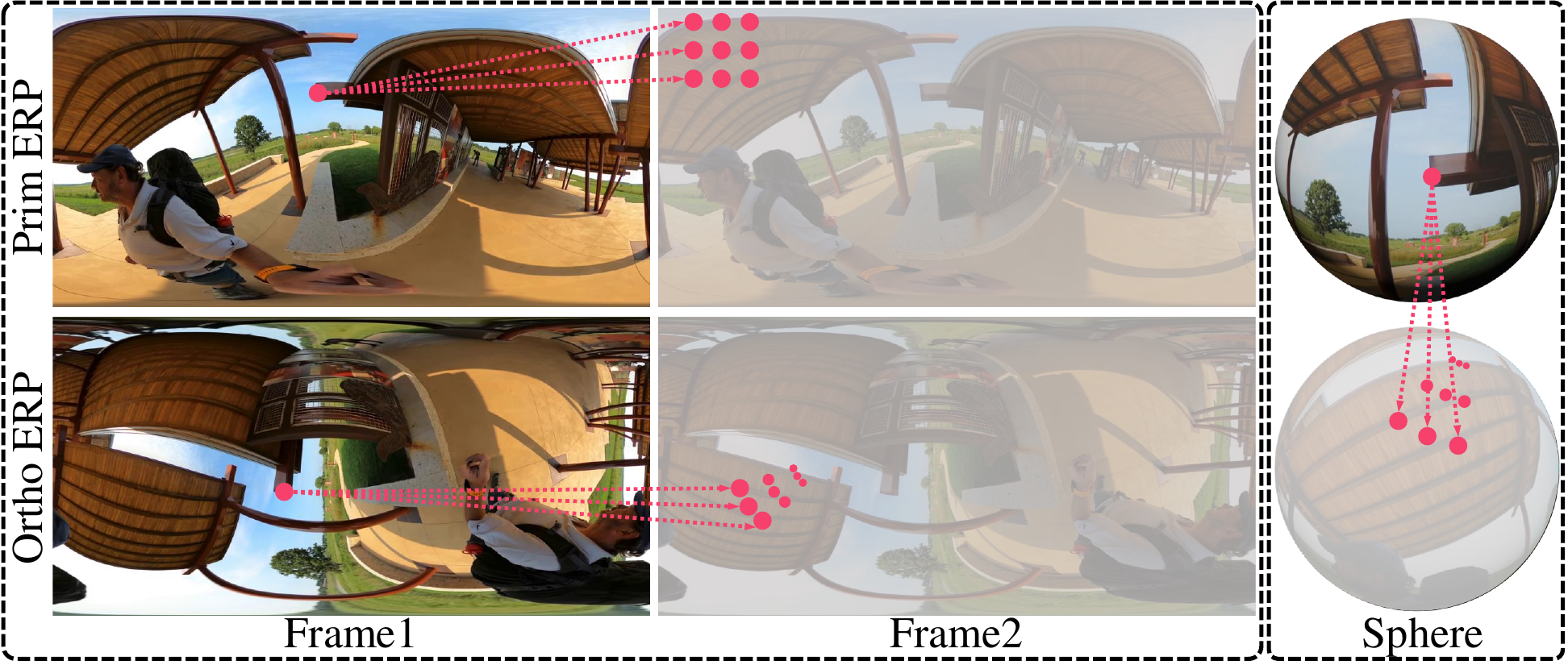}
    \vspace{-20pt}
    \caption{Schematic diagram of the Dual-Cost Collaborative Lookup (DCCL) operator.}
    \label{fig:dccl}
    \vspace{-15pt}
\end{figure}
%%%%%%%%%%%%%%%%%%%%%%%%%%%%%%%%%%%%%%%%%%%%%%%%%%%%%%%%%%

As shown in Fig.~\ref{fig:dccl}, DCCL performs correlation lookup on a unified spherical surface, mapping the indexed points to both the primitive image ${I}^p_2$ and the orthogonal image ${I}^o_2$, followed by separate lookups in $\{C^p_i\}$ and $\{C^o_i\}$. Given a current estimate of the primitive optical flow ${\mathcal{F}^p} = \{f^p_1, f^p_2\}$, we map each pixel $\boldsymbol{\mathrm{x}}^p = \{u^p, v^p\}$ in ${I}^p_1$ to its estimated correspondence in ${I}^p_2$:
\begin{equation}
    \boldsymbol{\hat{\mathrm{x}}}^p = \{(u^p + {f}^p_1(\boldsymbol{\mathrm{x}}^p))\bmod W, v^p + f^p_2(\boldsymbol{\mathrm{x}}^p)\}
\end{equation}
where the modulo operation explicitly enforces horizontal boundary continuity for panoramic images. We then define a local grid around $\boldsymbol{\hat{\mathrm{x}}}^p$:
\begin{equation}
    \mathcal{N}(\boldsymbol{\hat{\mathrm{x}}}^p)_r^p = \left\{ \boldsymbol{\hat{\mathrm{x}}}^p + \boldsymbol{\mathrm{dx}} \,
    | \, \boldsymbol{\mathrm{dx}} \in \mathbb{Z}^2, \, \|\boldsymbol{\mathrm{dx}}\|_1 \leq r \right\}
\end{equation}
We use $\mathcal{N}(\boldsymbol{\hat{\mathrm{x}}}^p)_r^p$ to index the primitive correlation cues $\mathcal{C}^p$ from $\{C^p_i\}$. Then we convert the primitive local grid $\mathcal{N}(\boldsymbol{\hat{\mathrm{x}}}^p)_r$ into the orthogonal local grid:
\begin{equation}
    \mathcal{N}(\boldsymbol{\hat{\mathrm{x}}}^p)_r^o = \left\{ \mathcal{R}(90^{\circ}, \boldsymbol{\mathrm{x}}) \, | \, \boldsymbol{\mathrm{x}} \in \mathcal{N}(\boldsymbol{\hat{\mathrm{x}}}^p)_r^p \right\}
\end{equation}
We use $\mathcal{N}(\boldsymbol{\hat{\mathrm{x}}}^p)_r^o$ to index the orthogonal correlation cues $\mathcal{C}^o$ from $\{C^o_i\}$. Then, we convert the cost map back to the primitive format:
\begin{equation}
    \mathcal{C}^{o2p} = T_o^p(\mathcal{C}^o)
\end{equation}

%%%%%%%%%%%%%%%%%%%%% MPF 可视化对比 %%%%%%%%%%%%%%%%%%%%%%%%%%%%%%%%%%%%%%%%
\begin{figure*}[ht]
    \centering
    \includegraphics[width=0.95\linewidth]{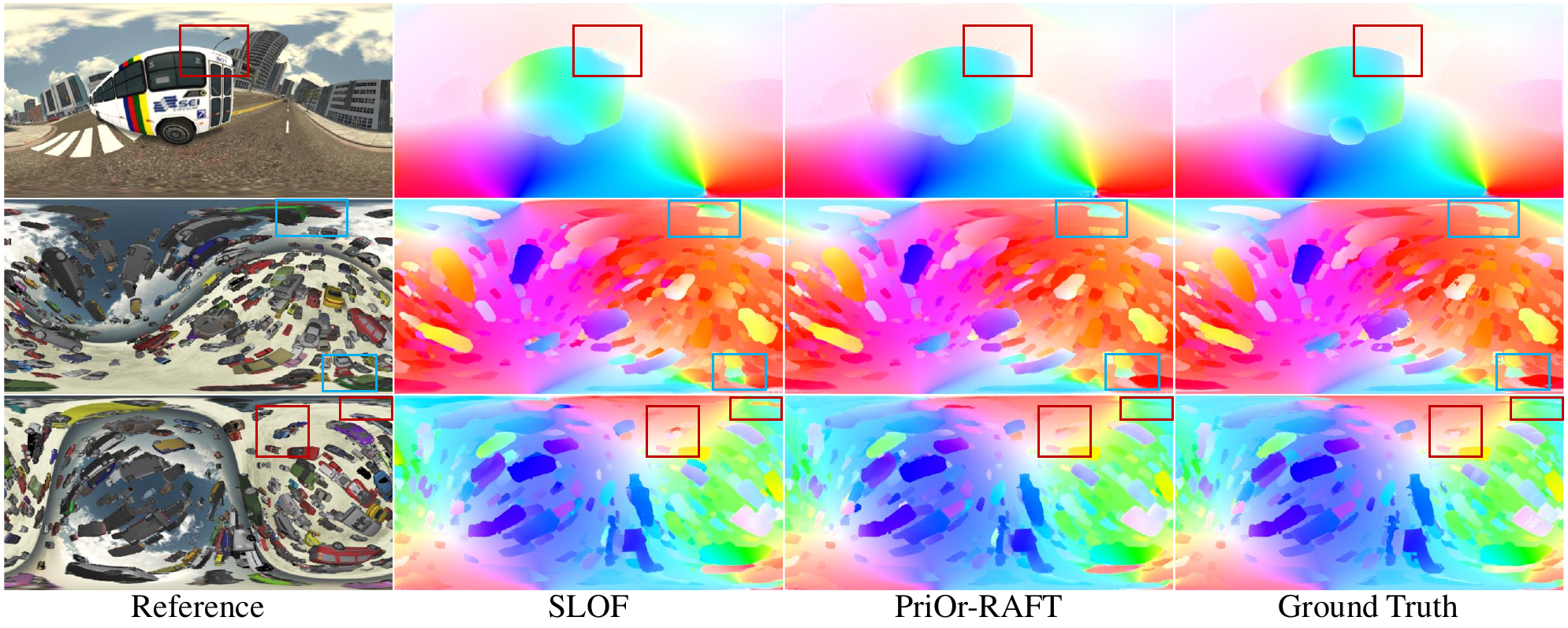}
    \vspace{-10pt}
    \caption{Qualitative results on the test set of MPFDataset. Our PriOr-RAFT outperforms SLOF in polar regions.}
    \label{fig:vis_mpf}
    \vspace{-15pt}
\end{figure*}
%%%%%%%%%%%%%%%%%%%%%%%%%%%%%%%%%%%%%%%%%%%%%%%%%%%%%%%%%%%%%%%%%%%%%%%%%%%%

$\mathcal{C}^p$ and $\mathcal{C}^{o2p}$ will be fed into the confidence-guided ConvGRU to jointly guide the restoration of the primitive flow field. The correlation information indexed from the orthogonal cost volume can effectively suppress the distortion noise in the polar regions of the primitive cost volume, compensating for the optical flow recovery in these regions.

\subsection{Orthogonal Branch}
\label{sec:orthobranch}
Given two consecutive frames $\{I^p_1, I^p_2\}$, referred to as the primitive view, we first convert both to the orthogonal view:
\begin{equation}
    \{I^o_1, I^o_2\} \; = \; \{T_p^o(I^p_1), T_p^o(I^p_2)\}
\end{equation}
Then, following RAFT \cite{raft}, we input $\{I^o_1, I^o_2\}$ into the feature network to extract the corresponding feature maps $\{\mathrm{f}^o_1, \mathrm{f}^o_2\}$, while feeding $I^o_1$ into the context network to obtain contextual features. After that, we use $\{\mathrm{f}^o_1, \mathrm{f}^o_2\}$ to construct the orthogonal cost volume:
\begin{equation}
    {C}^o_{ijkl} \; = \; \sum_h(f^o_1)_{hij} \cdot (f^o_2)_{hkl}
    \label{equ:corr}
\end{equation}
We then construct a 4-level correlation pyramid $\{C^o_i\}(i=1,2,3,4)$ by using 2D average pooling with a kernel size of 2 and a stride of 2 as the last two dimensions. Finally, we use a ConvGRU to iteratively update the optical flow field. In each iteration, we use the current optical flow estimation $\mathcal{F}^o$ to retrieve correlation information from both the orthogonal correlation pyramid $\{C^o_i\}$ and the primitive correlation pyramid $\{C^p_i\}$:
\begin{equation}
    \mathcal{C}^o, \mathcal{C}^{p2o} \; = \; \text{DCCL}(\{C^o_i\}, \{C^p_i\}, \mathcal{F}^o)
\end{equation}
$\mathcal{C}^o$ and $\mathcal{C}^{p2o}$ will be fed into the ConvGRU to predict the residual flow $\Delta f^o$ for updating:
 
\begin{equation}
    \hat{\mathcal{F}}^o = \mathcal{F}^o + \Delta f^o
\end{equation}

%%%%%%%%%%%%%%%%%%%%%%%%%%%%%%%%%%%%%%%%%%%%%%%%%%%%%%%%%%%%%%%%%%%%%%%%%%%%
%%%%%%%%%%%%                         Primitive Branch                %%%%%%%%%%%%%%%%
\subsection{Primitive Branch}
\label{sec:primbranch}

Symmetric to the orthogonal branch, the primitive branch first feeds the primitive frames $\{I^p_1, I^p_2\}$ into the encoder to obtain the feature maps $\{\mathrm{f}^p_1, \mathrm{f}^p_2\}$ and constructs the correlation pyramid $\{C^p_i\}(i=1,2,3,4)$ according to Eq.~\eqref{equ:corr}.

\begin{figure*}[ht]
    \centering
    \includegraphics[width=0.95\linewidth]{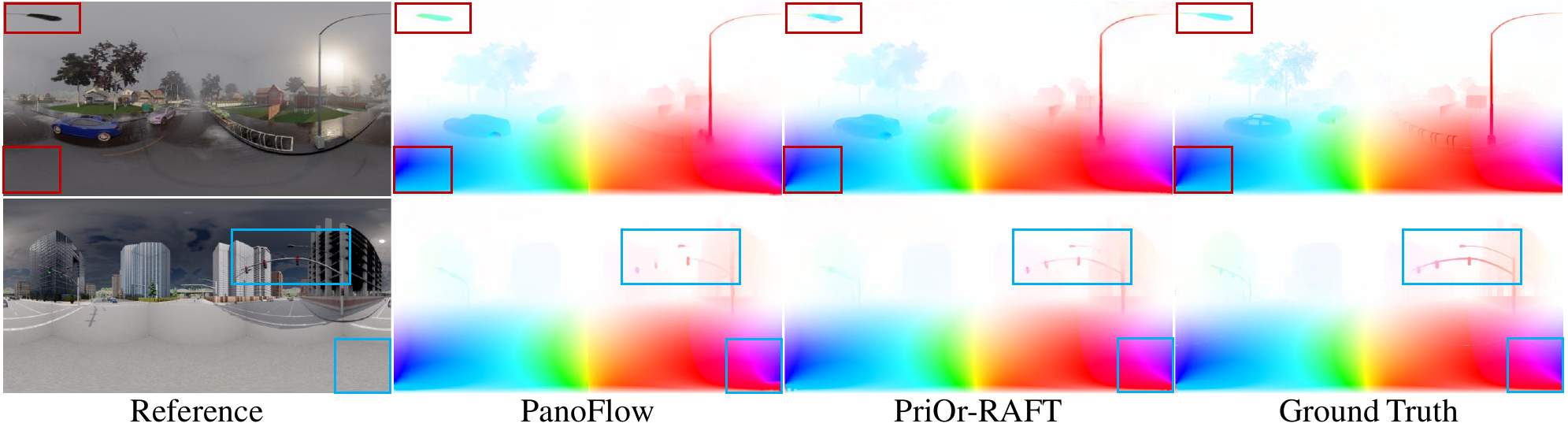}
    \vspace{-10pt}
    \caption{Qualitative results on the test set of FlowScape. Our PriOr-RAFT outperforms PanoFlow in the polar regions as well as in handling small horizontal objects.}
    \label{fig:vis_flowscape}
    \vspace{-18pt}
\end{figure*}

\textbf{Ortho-Driven Distortion Compensation (ODDC)} To further mitigate the effects of polar distortion, ODDC leverages the low-distortion priors of the orthogonal view to compensate for the flow field reconstruction in the polar regions of the primitive view. To explicitly represent the advantages of each view in different regions, we compute confidence maps of the primitive flow and the orthogonal flow, respectively. Specifically, during each iteration, given the current flows $\mathcal{F}^p$ and $\mathcal{F}^o$, we first convert the orthogonal flow $\mathcal{F}^o$ to primitive format:
\begin{equation}
    \mathcal{F}^{o2p} \; = \; T_o^p(\mathcal{F}^o)
\end{equation}

Then we compute both the primitive and the orthogonal confidence by computing the group-wise correlation following \cite{gwcnet, crestereo}:
\begin{equation}
    \begin{aligned}
        G^p(\boldsymbol{\mathrm{x}}) \; =& \; GW(\mathrm{f}^p_1(\boldsymbol{\mathrm{x}}), \mathrm{f}^p_2(\boldsymbol{\mathrm{x}} + \mathcal{F}^p) \bmod W )) \\
        G^{o2p}(\boldsymbol{\mathrm{x}}) \; =& \; GW(\mathrm{f}^p_1(\boldsymbol{\mathrm{x}}), \mathrm{f}^p_2(\boldsymbol{\mathrm{x}} + \mathcal{F}^{o2p}) \bmod W ))
    \end{aligned}
\end{equation}
where $GW$ denotes the group-wise correlation operation. In addition to the confidence information, we also use the current primitive flow $\mathcal{F}^p$ to index from the primitive cost volume and the orthogonal cost volume to obtain the correlation cues to further guide the restoration of the flow field:
\begin{equation}
    \mathcal{C}^p, \mathcal{C}^{o2p} \; = \; DCCL(\{{C}^p_i\}, \{{C}^o_i\}, \mathcal{F}^p) 
\end{equation}
After obtaining the correlation cues, confidence information, and flow conditions, these inputs are fed into a confidence-guided motion encoder that aggregates the primitive motion feature $m^p$ with compensation from the orthogonal view:
\begin{equation}
    \begin{aligned}
    m^p \; =& \; [\mathrm{En}_c(\mathcal{C}^p + \mathcal{C}^{o2p}), \mathrm{En}_g([G^p, G^{o2p}]), \\
    & \; \mathrm{En}_f([\mathcal{F}^p, \mathcal{F}^{o2p}]), \mathcal{F}^p, \mathcal{F}^{o2p}]
    \end{aligned}
\end{equation}

Here, $\mathrm{En}_c$, $\mathrm{En_{g}}$ and $\mathrm{En}_f$ are shallow convolutional layers designed to encode motion features. The aggregated feature $m^p$ is then passed to a ConvGRU to update the hidden state of the primitive branch. Subsequently, two convolutional layers decode a residual flow $\Delta f^p$, and the primitive flow is updated in accordance with RAFT:
\begin{equation}
    \hat{\mathcal{F}}^p = \mathcal{F}^p + \Delta f^p
\end{equation}

% \subsection{Loss Function}
% \label{sec:loss}
% To adapt to the non-uniform sampling characteristics of ERP projection on spherical surfaces, we weight the L1 loss of each pixel by its corresponding spherical area \cite{zhang2018saliency, MPF-net}:
% \begin{equation}
%     \Vert \mathcal{F}_{pr} - \mathcal{F}_{gt}\Vert_1^{sph} \; = \; \frac{1}{\sum \omega^j}\sum_{j\in \Omega}\Vert \mathcal{F}_{pr}^j - \mathcal{F}_{gt}^j \Vert_1 \cdot \omega ^j
% \end{equation}
% We supervise the outputs of the primitive branch and orthogonal branch using ground truth of the primitive view and orthogonal view respectively and follow \cite{raft} to exponentially increase the weights as the number of iterations increases. The total loss $\mathcal{L}$ is defined as the sum of the primitive branch loss $\mathcal{L}_p$ and the orthogonal branch loss $\mathcal{L}_o$ as follows:
% \begin{equation}
%     \begin{aligned}
%         \mathcal{L}_p \; =& \; \sum_{i=1}^{N}\gamma^{N-i}\Vert \mathcal{F}^p_i - \mathcal{F}^{gt} \Vert_1^{sph} \\
%         \mathcal{L}_o \; =& \; \sum_{i=1}^{N}\gamma^{N-i}\Vert \mathcal{F}^o_i - T_p^o(\mathcal{F}^{gt}) \Vert_1^{sph} \\
%         \mathcal{L} \; =& \; \mathcal{L}_p + \mathcal{L}_o
%     \end{aligned}
% \end{equation}
% where $\gamma = 0.8$, and $\mathcal{F}^{gt}$ is the ground truth of the primitive view.

\subsection{Loss Function}
\label{sec:loss}

To account for the non-uniform sampling characteristics of the ERP projection on spherical surfaces, we weight the L1 loss of each pixel by its corresponding spherical area \cite{zhang2018saliency, MPF-net}. This is expressed as:
\begin{equation}
    \Vert \mathcal{F}_{pr} - \mathcal{F}_{gt} \Vert_1^{sph} \; = \; \frac{1}{\sum \omega^j} \sum_{j \in \Omega} \Vert \mathcal{F}_{pr}^j - \mathcal{F}_{gt}^j \Vert_1 \cdot \omega^j
\end{equation}
where $\omega^j$ denotes the spherical area weight corresponding to each pixel.

For supervision, we use the ground truth of the primitive and orthogonal views to guide the outputs of the respective branches. The loss function follows the approach in \cite{raft}, where the weights are exponentially increased as the number of iterations progresses. The total loss $\mathcal{L}$ is the sum of the primitive branch loss $\mathcal{L}_p$ and the orthogonal branch loss $\mathcal{L}_o$:
\begin{equation}
    \begin{aligned}
        \mathcal{L}_p \; &= \; \sum_{i=1}^{N} \gamma^{N-i} \Vert \mathcal{F}^p_i - \mathcal{F}^{gt} \Vert_1^{sph} \\
        \mathcal{L}_o \; &= \; \sum_{i=1}^{N} \gamma^{N-i} \Vert \mathcal{F}^o_i - T_p^o(\mathcal{F}^{gt}) \Vert_1^{sph} \\
        \mathcal{L} \; &= \; \mathcal{L}_p + \mathcal{L}_o
    \end{aligned}
\end{equation}
where $\gamma = 0.8$ and $\mathcal{F}^{gt}$ represents the ground truth of the primitive view.

\section{Experiment}
\label{sec:experiment}

\subsection{Datasets \& Evaluation Metrics}
\textbf{Datasets.} We evaluate our model on both synthetic and real-world datasets. \textit{MPFDataset}~\cite{MPF-net} provides dense optical flow ground truth with two scenes: City (2000 training pairs, 138 testing pairs) and EFT (2211 training pairs, 99 testing pairs). \textit{FlowScape}~\cite{panoflow} features four weather scenarios with 5000 training pairs and 1400 testing pairs. Additionally, we test on real-world datasets \textit{OmniPhotos}~\cite{omniphotos} and \textit{ODVista}~\cite{odvista}, which contain panoramic videos but lack ground truth.

\textbf{Evaluation Metrics.} Following~\cite{TanImg, MPF-net}, we use end-point error (EPE) and spherical end-point error (SEPE) to evaluate panoramic optical flow:
\begin{equation}
    SEPE \; = \; \frac{1}{|\Omega|}\sum_{i\in \Omega} d(\mathcal{F}^{pr}_i, \mathcal{F}^{gt})
\end{equation}
where $d(\cdot, \cdot)$ denotes the geodesic distance on the unit sphere.

\subsection{Implementation Details}
\label{sec:implementation}
We implement PriOr-Flow using PyTorch and conduct experiments on NVIDIA RTX 3090 GPUs. Following the baseline approach in RAFT~\cite{raft}, we use the AdamW~\cite{adamw} optimizer and clip gradients within the range of [-1, 1]. The learning rate is scheduled using a one-cycle policy, with an initial learning rate of 1e-4. Model weights are initialized with the pre-trained RAFT weights on FlyingThings~\cite{flyingthings} before fine-tuning on the omnidirectional datasets. For the MPF-dataset, we use a batch size of 4, a learning rate of 1e-4, and train the model for 60k steps. For FlowScape, following PanoFlow~\cite{panoflow}, we use a batch size of 6, a learning rate of 1e-4, and train for 100k steps. During both training and testing, we set the number of iterations to 12.

\begin{table*}[t]\small
\centering
\renewcommand{\arraystretch}{1} % 调整行距
\setlength{\tabcolsep}{3.5pt} % 调整列间距
\begin{tabular}{l|ccc|cc|cc|cc}
\hline
\multirow{2}{*}{Model} &
  Orthogonal &
  \multirow{2}{*}{DCCL} &
  \multirow{2}{*}{ODDC} &
  \multicolumn{2}{c|}{\textbf{Equator}} &
  \multicolumn{2}{c|}{\textbf{Poles}} &
  \multicolumn{2}{c}{\textbf{All}} \\ \cline{5-10}
\multicolumn{1}{l|}{} &
  View &
      &
      &
  EPE &
  \multicolumn{1}{c|}{SEPE} &
  EPE &
  \multicolumn{1}{c|}{SEPE} &
  EPE &
  SEPE \\ \hline
Baseline (RAFT)            &  &  &  & 1.09 & 6.29 & 7.90 & 8.56 & 4.49 & 7.43  \\
Ortho + D               & \checkmark & \checkmark &  & 1.08 & 6.24 & 7.56 & 8.16 & 4.32 & 7.20 \\
% \begin{tabular}[c]{@{}l@{}}Ortho + D + O(w/o conf)\end{tabular} & \checkmark & \checkmark & \checkmark &  & 1.06 & 6.14 & 5.65 & 6.60 & 3.36 & 6.37  \\
Full Model (PriOr-RAFT)    & \checkmark & \checkmark & \checkmark & \textbf{1.03} & \textbf{5.99} & \textbf{5.57} & \textbf{6.47} & \textbf{3.30} & \textbf{6.23}  \\ 
\bottomrule
\end{tabular}
\vspace{-8pt}
\caption{Ablation study of the effectiveness of proposed modules. DCCL stands for Dual-Cost Collaborative Lookup, and ODDC represents Ortho-Driven Distortion Compensation. We evaluate the performance of different models in both the equatorial and polar regions. Polar regions refer to areas with latitudes greater than $45^{\circ}$, while equatorial regions cover the rest of the sphere.}
\label{tab:ablation_modules}
\vspace{-10pt}
\end{table*}

%%%%%%%%%%%%%%%%%%%%%%%%%%%%%%%%%%%%%%%%%%%%%%%%%%%%%%%%%%%%%%%%%%%%%%%%%%%%%%%%%%%%%%%%%

%%%%%%%%%%%%%%%%%%%%%%%%%%%%%%%%%%%%%%%%% 真实场景可视化对比 %%%%%%%%%%%%%%%%%%%%%%%%%%%%%%%%%%%%
\begin{figure*}[ht]
    \centering
    \includegraphics[width=0.95\linewidth]{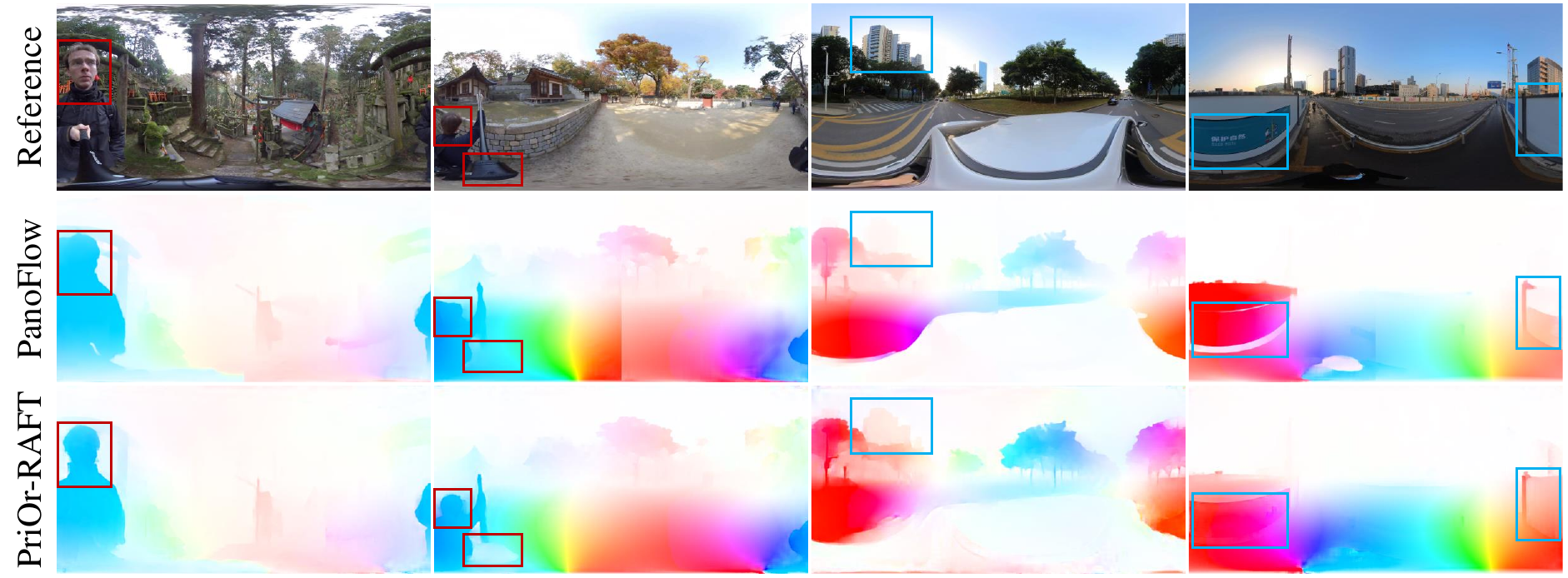}
    \vspace{-10pt}
    \caption{Qualitative results on real-world scenarios. The left two columns present the results on the OmniPhotos dataset, while the right two columns display the results on the ODVista dataset.}
    \label{fig:vis_real}
    \vspace{-15pt}
\end{figure*}
%%%%%%%%%%%%%%%%%%%%%%%%%%%%%%%%%%%%%%%%%%%%%%%%%%%%%%%%%%%%%%%%%%%%%%%%%%%%%%%%%%%%%%%%%

%%%%%%%%%%%%%%%%%%%%%%%%%%%%%%%%%%%%%%%%% 置信度表格 %%%%%%%%%%%%%%%%%%%%%%%%%%%%%%%%%%%%%%%%
\begin{table}[h]\small
    \centering
    \renewcommand{\arraystretch}{1}
    \setlength{\tabcolsep}{3.5pt}
        \begin{tabular}{@{}l|c|cc|cc@{}}
            \toprule
            \multirow{2}{*}{Model} & Confidence & \multicolumn{2}{c|}{Poles} & \multicolumn{2}{c}{All} \\
            \cline{3-6} 
            &   Type     &    EPE  &   SEPE           &     EPE  &    SEPE \\
            \midrule
            Ours w/o conf               & -          &    5.65 &   6.60           &     3.36 &    6.37 \\
            Ours dist conf          & Distortion Map &    5.62 &   6.55           &     3.34 &    6.34 \\
            Ours (PriOr-RAFT)       & Warp\&GW   & \textbf{5.57} & \textbf{6.47} & \textbf{3.30} & \textbf{6.23} \\
            \bottomrule
    \end{tabular}
%	}
    \vspace{-10pt}
    \caption{Ablation study on the type of confidence in ODDC.}
    \vspace{-22pt}
    \label{tab:conf}
\end{table}

\subsection{Ablation Study}
To evaluate the effectiveness of each component in PriOr-Flow, we conduct ablation experiments on the EFT scene of MPFDataset~\cite{MPF-net}, which contains diverse motion patterns.

% \textbf{Effectiveness of proposed modules.} We progressively incorporate our proposed modules into the baseline model and assess performance across different regions. As shown in Tab.~\ref{tab:ablation_modules}, compared to the original RAFT operation, our DCCL module effectively suppresses distortion noise in the cost volume, yielding notable improvements. Adding ODDC—even without incorporating confidence information—still leads to significant gains, primarily benefiting polar regions (an improvement of 25. 3\% in the EPE metric). This suggests that motion cues from the orthogonal branch can compensate for distortions in the primitive view. When ODDC is properly integrated with confidence guidance, the complete model (PriOr-RAFT) achieves the best performance, improving accuracy by 26.5\%.

\textbf{Effectiveness of proposed modules.} We progressively incorporate our proposed modules into the baseline model and assess performance across different regions. As shown in Tab.~\ref{tab:ablation_modules}, compared to the original RAFT operation, our DCCL module effectively suppresses distortion noise in the cost volume, yielding notable improvements. Adding ODDC leads to significant gains, primarily benefiting polar regions (an improvement of 26.3\% in the EPE metric). This suggests that motion cues from the orthogonal branch can compensate for distortions in the primitive view. We also perform an ablation study on the confidence guidance in ODDC, as shown in Tab.~\ref{tab:conf}. While the distortion map provides view-specific confidence, it is fixed and independent of the input images and optical flow, limiting its ability to reflect actual flow confidence. In contrast, our method explicitly measures feature similarity before and after warping, yielding more accurate confidence estimation and enabling more effective flow fusion.

\begin{table}[t]\small
\centering
\renewcommand{\arraystretch}{1}
\setlength{\tabcolsep}{4pt}
\begin{tabular}{lccccc}
\toprule
\multirow{2}{*}{Model} & \multicolumn{2}{c}{EPE}  & SEPE & Run-time \\ \cmidrule(lr){2-3} 
                       & (px) & Reduction &  (mm) & (s)  \\ \midrule
RAFT~\cite{raft}                   & 4.49  & -               & 7.43  & 0.07  \\
PriOr-RAFT (4-iter)    & 3.89  & \textcolor{red}{$\downarrow$ 13.4\%}   & 7.17  & 0.10  \\
PriOr-RAFT             & 3.30  & \textcolor{red}{$\downarrow$ 26.5\%}   & 6.23  & 0.20  \\ \hline
GMA~\cite{GMA}                    & 4.26  & -               & 7.07  & 0.07  \\
PriOr-GMA (4-iter)     & 3.51  & \textcolor{red}{$\downarrow$ 17.6\%}   & 6.68  & 0.10  \\
PriOr-GMA              & 3.25  & \textcolor{red}{$\downarrow$ 23.7\%}   & 6.17  & 0.20  \\ \hline
SKFlow~\cite{skflow}                 & 3.79  & -               & 6.55  & 0.12  \\
PriOr-SKFlow (4-iter)  & 3.33  & \textcolor{red}{$\downarrow$ 12.1\%}   & 6.28  & 0.13  \\
PriOr-SKFlow           & 3.19  & \textcolor{red}{$\downarrow$ 15.8\%}   & 6.13  & 0.30  \\ 
\bottomrule
\end{tabular}
\vspace{-10pt}
\caption{Ablation study of the universality of PriOr-Flow.}
\label{tab:universality}
\vspace{-10pt}
\end{table}

%%%%%%%%%%%%%%%%%%%%%%%%%%%%%%%% 正交视图的选择 %%%%%%%%%%%%%%%%%%%%%%%%%%%%%
\begin{table}[t]\small
    \centering
    \renewcommand{\arraystretch}{1}
    \setlength{\tabcolsep}{4pt}
    \begin{tabular}{lcccc}
        \toprule
        View & Axis & $\theta$ & EPE(px) & SEPE(mm) \\ 
        \midrule
        y-ortho & y    & $90^{\circ}$   & 3.55    & 6.46    \\
        x-semi-ortho & x    & $45^{\circ}$   & 3.41    & 6.46    \\
        x-ortho & x    & $90^{\circ}$  & \textbf{3.30}  & \textbf{6.23}  \\ 
        \bottomrule
    \end{tabular}
    \vspace{-10pt}
    \caption{Ablation study of the selection of the orthogonal view.}
    \label{tab:axis_results}
    \vspace{-10pt}
\end{table}
%%%%%%%%%%%%%%%%%%%%%%%%%%%%%%%%%%%%%%%%%%%%%%%%%%%%%%%%%%%%%%%%%%%%%%%%%

%%%%%%%%%%%%%%%%%%%%%%%%%%%%%%%%%%% 迭代次数分析 %%%%%%%%%%%%%%%%%%%%%%%%%%%%%%%%%%
\begin{table}[t]\small
    \centering
    \renewcommand{\arraystretch}{1}
    \setlength{\tabcolsep}{4pt}
    \begin{tabular}{c|l|ccccc}
        \toprule
        \multirow{2}{*}{Metric} & \multirow{2}{*}{Model} & \multicolumn{5}{c}{Number of Iterations} \\ 
        \cline{3-7}
        &  & 3 & 4 & 8 & 12 & 16 \\ 
        \hline
        \multirow{2}{*}{\shortstack{EPE \\ (px)}}
        & RAFT    & 6.09 & 5.49 & 4.64 & 4.49 & 4.48 \\
        & PriOr-RAFT  & 4.37 & 3.89 & 3.39 & 3.30 & 3.30 \\
        % \hline
        % \multirow{2}{*}{\shortstack{SEPE \\ (mm)}} 
        % & RAFT    & 9.57 & 8.71 & 7.61 & 7.43 & 7.41 \\
        % & PriOr-RAFT  & 7.99 & 7.17 & 6.35 & 6.23 & 6.22 \\
        \bottomrule
    \end{tabular}
    \vspace{-10pt}
    \caption{Ablation study of the number of iterations.}
    \label{tab:iterations}
    \vspace{-20pt}
\end{table}

%%%%%%%%%%%%%%%%%%%%%%%%%%%%%%%%%%%%%%%%%%%%%%%%%%%%%%%%%%%%%%%%%%%%%%%%%%%%%%%%

%%%%%%%%%%%%%%%%%%%%%%%%%%%%%%  和sota对比表格 %%%%%%%%%%%%%%%%%%%%%%%%%%%%%%%%%
% \usepackage{booktabs}
% \usepackage{multirow}
\begin{table*}[t]\small
\setlength{\tabcolsep}{2.5pt}
\centering
\begin{tabular}{l|l|cc|cc|cc|cc|cc|cc|cc|cc}
\toprule
\multirow{3}{*}{Method} &
  \multirow{3}{*}{Baseline} &
  \multicolumn{6}{c|}{MPFDataset \cite{MPF-net}} &
  \multicolumn{10}{c}{FlowScape \cite{panoflow}} \\ \cline{3-18} 
 &
   &
  \multicolumn{2}{c}{EFT} &
  \multicolumn{2}{c}{City} &
  \multicolumn{2}{c|}{All} &
  \multicolumn{2}{c}{Sunny} &
  \multicolumn{2}{c}{Cloud} &
  \multicolumn{2}{c}{Fog} &
  \multicolumn{2}{c}{Rain} &
  \multicolumn{2}{c}{All} \\ \cline{3-18} 
          &         & EPE  & SEPE  & EPE & SEPE & EPE  & SEPE & EPE  & SEPE & EPE  & SEPE & EPE  & SEPE & EPE  & SEPE & EPE  & SEPE\\ \hline
SphereNet~\cite{spherenet} & RAFT~\cite{raft}    & 13.2 & 15.7  & 8.28 & 7.44 & 10.7 & 19.4 & 13.7 & 22.9 & 9.94 & 15.6 & 12.5 & 19.3 & 13.3 & 20.2 & 12.9 & 21.0 \\
TanImg~\cite{TanImg}    & DIS~\cite{DIS}     & 8.04 & 19.3  & 3.74 & 6.48 & 5.89 & 12.9 & 19.6 & 31.1 & 15.6 & 25.4 & 17.6 & 27.8 & 19.6 & 31.9 & 18.7 & 29.9 \\
TanImg~\cite{TanImg}    & RAFT~\cite{raft}    & \cellcolor{yellow!25} 4.38 & 9.52  & 3.13 & 5.06 & 3.76 & 7.29 & 19.6 & 28.2 & 14.7 & 17.9 & 17.8 & 23.8 & 17.3 & 22.7 & 18.3 & 25.3 \\
MPF-net~\cite{MPF-net}   & PWC-net~\cite{pwc-net} & 5.06 & 10.49 & 1.78 & 3.24 & 3.42 & 6.87 & -  &  -    &    -  &    -  &    -  &    -  &    -  &  - & -  &   -  \\
SLOF~\cite{slof}      & RAFT~\cite{raft}    & 4.98 & \cellcolor{yellow!25} 8.20  & \cellcolor{yellow!25}1.35 & \cellcolor{yellow!25}2.06 & \cellcolor{yellow!25}3.17 & \cellcolor{yellow!25}5.13 &  7.84 & 6.07  & 6.00  &  3.64 & 8.08  & 6.35  & 7.66 &  6.24 & 7.59 & 5.79 \\
PanoFlow~\cite{panoflow}  & RAFT~\cite{raft}    &   -   &    -   &    -  &   - &  - &  - & 3.61 & 5.19 & \cellcolor{yellow!25}1.38 & 1.97 & 3.60 & 4.78 & 4.25 & 5.92 & 3.38 & 4.78 \\
PanoFlow~\cite{panoflow} & CSFlow~\cite{csflow} & -   &   -  &  - &  -  & -  & -  &  
 \cellcolor{yellow!25}3.55  &   \cellcolor{yellow!25}4.76   &   1.47   &   \cellcolor{yellow!25}1.86   &  \cellcolor{yellow!25} 3.55   & \cellcolor{yellow!25} \cellcolor{yellow!25}4.70    &   \cellcolor{yellow!25}3.93   & \cellcolor{yellow!25}5.46  & \cellcolor{yellow!25}3.31 &   \cellcolor{yellow!25}4.44  \\
PriOr-Flow & RAFT~\cite{raft}    & \cellcolor{red!25}\textbf{3.30} & \cellcolor{red!25}\textbf{6.23}  & \cellcolor{red!25}\textbf{1.13} & \cellcolor{red!25}\textbf{1.88} & \cellcolor{red!25}\textbf{2.22} & \cellcolor{red!25}\textbf{4.06} & \cellcolor{red!25}\textbf{2.41} & \cellcolor{red!25}\textbf{3.56} & \cellcolor{red!25}\textbf{0.93} & \cellcolor{red!25}\textbf{1.36} & \cellcolor{red!25}\textbf{2.81} & \cellcolor{red!25}\textbf{4.09} & \cellcolor{red!25}\textbf{2.95} & \cellcolor{red!25}\textbf{4.71} & \cellcolor{red!25}\textbf{2.33} & \cellcolor{red!25}\textbf{3.49} \\ \bottomrule
\end{tabular}
\vspace{-10pt}
\caption{Quantitative evaluation on MPFDataset and FlowScape. The \colorbox{red!25}{\textbf{best}} and \colorbox{yellow!25}{second best} are marked with colors. ``-'' denotes that the model weights are not publicly available.}
\label{tab:sota}
\vspace{-18pt}
\end{table*}
%%%%%%%%%%%%%%%%%%%%%%%%%%%%%%%%%%%%%%%%%%%%%%%%%%%%%%%%%%%%%%%%%%%%%%%%%%%%%%%%%%

%%%%%%%%%%%%%%%%%%%%%%%%%%%%%%%%%%% 和sota的两极区域精度对比 %%%%%%%%%%%%%%%%%%%%%%%%%%
\begin{table}[t]\small
\centering
\renewcommand{\arraystretch}{1} % 增加行距，提高可读性
\setlength{\tabcolsep}{5pt} % 调整列间距
\begin{tabular}{l|cc|cc|cc}
\toprule
\multirow{2}{*}{Model} & \multicolumn{2}{c|}{Equator} & \multicolumn{2}{c|}{Poles} & \multicolumn{2}{c}{All} \\ \cline{2-7}
   & EPE  & SEPE & EPE  & SEPE & EPE  & SEPE \\  
\hline
PanoFlow   & 0.52 & 2.87 & 6.25 & 6.68 & 3.38 & 4.78 \\
PriOr-RAFT & 0.53 & 2.94 & 4.13 & 4.03 & 2.33 & 3.49 \\ 
\bottomrule
\end{tabular}
\vspace{-8pt}
\caption{Quantitative evaluation on FlowScape test set in different regions.}
\label{tab:polar_compare}
\vspace{-20pt}
\end{table}
%%%%%%%%%%%%%%%%%%%%%%%%%%%%%%%%%%%%%%%%%%%%%%%%%%%%%%%%%%%%%%%%%%%%%%%%%%%%%%%%%%%%%%%%%

% \textbf{Universality of proposed modules.} To validate the universality of the proposed modules, we employ three classic iterative perspective optical flow estimation methods as baselines. We replace the original cost volume lookup with DCCL and introduce an additional orthogonal branch, while substituting the original ConvGRU with ODDC in the primitive branch. As shown in Tab.~\ref{tab:universality}, all methods achieved significant improvements in both the EPE and SEPE metrics. As shown in Tab.~\ref{tab:universality}, all methods achieved significant improvements in both the EPE and SEPE metrics. Moreover, when we reduced the number of iterations to 4, these methods still managed to achieve substantial performance gains while keeping the inference time nearly unchanged.

\textbf{Universality of proposed modules.} To verify the general applicability of our modules, we integrate them into three classic iterative perspective optical flow estimation methods as baselines. Specifically, we replace the original cost volume lookup with DCCL, introduce an additional orthogonal branch, and substitute the primitive branch's ConvGRU with ODDC. As shown in Tab.~\ref{tab:universality}, all methods achieve significant improvements in both EPE and SEPE. Furthermore, even when reducing the number of iterations to 4, these methods still achieve substantial accuracy gains while maintaining nearly the same inference time. In particular, PriOr-GMA accomplishes a 17.6\% performance improvement with almost identical inference time.

% \textbf{The selection of the orthogonal view}. We verify different rotation axes and various rotation angles $\theta$ as shown in Tab.~\ref{tab:axis_results}. When we rotate the spherical image $90^{\circ}$ around the y-axis, the polar region is split into two parts located on the left and right sides of the ERP image. The disruption of regional continuity makes it more challenging for the model to learn orthogonal optical flow. When we rotate the spherical image $45^{\circ}$ around the x-axis, the low-distortion region of the ``semi-orthogonal'' view does not cover as much of the primitive image as a $90^{\circ}$ rotation does, resulting in suboptimal compensation for the polar regions in the original branch. Consequently, we selected the image rotated $90^{\circ}$ around the x-axis as the orthogonal view for PriOr-Flow.

% \textbf{The selection of the orthogonal view}. We evaluate different rotation axes and angles $\theta$ as shown in Tab.~\ref{tab:axis_results}. Rotating the spherical image by $90^{\circ}$ around the y-axis splits the polar region into two separate parts on the left and right of the ERP image, disrupting regional continuity and making orthogonal optical flow estimation more challenging. A $45^{\circ}$ rotation around the x-axis creates a ``semi-orthogonal'' view, but its low-distortion region covers less of the primitive image compared to a $90^{\circ}$ rotation, leading to suboptimal compensation for polar regions. Based on these findings, we choose the $90^{\circ}$ x-axis rotation as the orthogonal view for PriOr-Flow.

\textbf{The selection of the orthogonal view.} We evaluate different rotation axes and angles $\theta$ as shown in Tab.~\ref{tab:axis_results}. A $90^{\circ}$ rotation around the y-axis splits the polar region, disrupting continuity and complicating orthogonal flow estimation. A $45^{\circ}$ rotation around the x-axis results in a "semi-orthogonal" view, but its low-distortion region covers less of the primitive image, leading to suboptimal compensation for polar regions. Therefore, we select the $90^{\circ}$ x-axis rotation as the orthogonal view for PriOr-Flow.

% \textbf{Number of iterations and efficiency.} Our PriOr-Flow can achieve better performance with a smaller number of iterations. As shown in Tab.~\ref{tab:iterations}, compared to the baseline, our PriOr-RAFT achieves better performance with only 3 iterations under the same inference time. This demonstrates that introducing motion features from the orthogonal branch in the original branch can significantly accelerate the convergence speed of the model.

\textbf{Number of iterations.} PriOr-Flow achieves better performance with fewer iterations. As shown in Tab.~\ref{tab:iterations}, compared to the baseline, PriOr-RAFT outperforms it with just 3 iterations. This demonstrates that incorporating motion features from the orthogonal branch significantly accelerates model convergence.

\vspace{-8pt}
% \subsection{Comparisons with State-of-the-art} We quantitatively compare our Prior-RAFT with other panoramic optical flow estimation methods, including weight transformation-based methods, tangent plane-based methods~\cite{TanImg}, and ERP-based methods~\cite{MPF-net, slof, panoflow}, on the MPFDataset~\cite{MPF-net} and FlowScape~\cite{panoflow} datasets. For the weight transformation-based methods, we opt to employ the more efficient and rapid weight transformation technique from SphereNet~\cite{spherenet} applied to RAFT, which is pre-trained on FlyingThings~\cite{flyingthings}, instead of using equirectangular convolution~\cite{equiconv} as in OmniFlowNet~\cite{omniflownet}. We discover that this approach yields superior performance, achieving an EPE metric of 12.9 on FlowScape, compared to the 19.6 EPE of the OmniFlowNet version. For the tangent plane-based methods, to ensure a fair comparison, we replace the DIS~\cite{DIS} originally used in \cite{TanImg} with RAFT, which is also pre-trained on FlyingThings. The results are shown in Tab.~\ref{tab:sota}.
\subsection{Comparisons with State-of-the-art}
\vspace{-5pt}
We quantitatively compare PriOr-RAFT with state-of-the-art panoramic optical flow estimation methods, including weight transformation-based, tangent plane-based~\cite{TanImg}, and ERP-based methods~\cite{MPF-net, slof, panoflow}, on MPFDataset~\cite{MPF-net} and FlowScape~\cite{panoflow}. For weight transformation-based methods, we adopt the more efficient SphereNet~\cite{spherenet} weight transformation applied to RAFT, pre-trained on FlyingThings~\cite{flyingthings}, instead of using equirectangular convolution~\cite{equiconv} as in OmniFlowNet~\cite{omniflownet}. This approach achieves an EPE of 12.9 on FlowScape, significantly outperforming OmniFlowNet’s 19.6 EPE. For tangent plane-based methods, to ensure fairness, we replace the original DIS~\cite{DIS} used in \cite{TanImg} with RAFT, also pre-trained on FlyingThings. The results are summarized in Tab.~\ref{tab:sota}.

% \textbf{MPFDataset.} Our PriOr-RAFT achieves state-of-the-art (SOTA) performance in both the EFT and City scenarios. Specifically, for the planar optical flow metric EPE, our method demonstrates a 30.0\% performance improvement compared to SLOF~\cite{slof}. For the spherical optical flow metric SEPE, our method achieves a 20.9\% performance improvement over SLOF~\cite{slof}. Fig.~\ref{fig:vis_mpf} presents a visual comparison between our method and MPF-net~\cite{MPF-net}.

% \textbf{FlowScape.} Our PriOr-RAFT achieves the best performance across all four weather scenarios. Overall, compared to the previous SOTA, we achieve a 29.6\% improvement in EPE and a 21.4\% improvement in SEPE. To validate the ability of our PriOr-Flow to compensate for the optical flow in the polar regions by leveraging the low-distortion prior of orthogonal views, we validate the performance of our method and PanoFlow in both the polar and equatorial regions. As shown in Tab.~\ref{tab:polar_compare}, our PriOr-RAFT outperforms PanoFlow by 39.7\% in the polar regions. However, since the low distortion in the polar regions of the orthogonal view comes at the cost of higher distortion in the equatorial regions, the introduction of orthogonal motion features introduces some interference to the optical flow in the equatorial regions. This results in our method being slightly inferior to PanoFlow in the equatorial regions. Fig.~\ref{fig:vis_flowscape} presents a visual comparison between PriOr-RAFT and PanoFlow (RAFT) on the FlowScape dataset.

\textbf{MPFDataset.} Our PriOr-RAFT achieves state-of-the-art performance in both the EFT and City scenarios. Specifically, for planar optical flow (EPE), our method surpasses SLOF~\cite{slof} by 30.0\%, while for spherical optical flow (SEPE), it achieves a 20.9\% improvement. Fig.~\ref{fig:vis_mpf} provides a visual comparison with SLOF~\cite{slof}.

\textbf{FlowScape.} PriOr-RAFT achieves the best performance across all weather conditions, improving EPE by 29.6\% and SEPE by 21.4\% over the previous SOTA. To verify its effectiveness in polar regions, we compare it with PanoFlow in different regions. As shown in Tab.~\ref{tab:polar_compare}, PriOr-RAFT outperforms PanoFlow by 39.7\% in polar regions. However, due to the trade-off between distortion in different areas, our method is slightly inferior to PanoFlow in the equatorial regions. Fig.~\ref{fig:vis_flowscape} presents a visual comparison between PriOr-RAFT and PanoFlow (RAFT) on FlowScape.
\vspace{-5pt}

% \subsection{Generalization to Real-World Scenarios.} Due to the lack of panoramic datasets with ground truth optical flow in real-world scenarios, to evaluate the performance of our method in such settings, we conducted qualitative comparisons of our PriOr-RAFT and PanoFlow~\cite{panoflow} on the OmniPhotos~\cite{omniphotos} and ODVista~\cite{odvista} datasets. Both models were trained on the FlowScape dataset and directly applied to these real-world datasets. The qualitative results are illustrated in Fig.~\ref{fig:vis_real}, which showcases the generalization capability of our approach in real-world environments.The ground region in the second column of Fig.~\ref{fig:vis_real} and the roof region in the third column demonstrate that our method is capable of effectively predicting optical flow in the polar regions.

\subsection{Generalization to Real-World Scenarios.} 
\vspace{-5pt}
Due to the lack of panoramic datasets with ground truth optical flow in real-world scenarios, we evaluate our method qualitatively on OmniPhotos~\cite{omniphotos} and ODVista~\cite{odvista} datasets. Both PriOr-RAFT and PanoFlow~\cite{panoflow} were trained on FlowScape and directly applied to these real-world datasets. Fig.~\ref{fig:vis_real} illustrates the qualitative results, demonstrating the generalization capability of our approach. Notably, the ground region in the second column and the roof region in the third column highlight our method’s ability to effectively predict optical flow in polar regions.
\vspace{-8pt}

\section{Conclusion}
\vspace{-5pt}
% \begin{sloppypar}
We propose PriOr-Flow, a novel and universal dual-branch method for panoramic optical flow estimation. By introducing the Dual-Cost Collaborative Lookup (DCCL) operator and the Ortho-Driven Distortion Compensation (ODDC) module, PriOr-Flow leverages the low-distortion prior from the orthogonal view to significantly improve performance, especially in polar regions. Our PriOr-RAFT model achieves state-of-the-art performance on both the MPFDataset and FlowScape, with notable improvements in accuracy. Additionally, PriOr-Flow demonstrates strong generalization in real-world scenarios, validating its effectiveness beyond synthetic datasets.
% \end{sloppypar}

% \begin{sloppypar}
\noindent\textbf{Acknowledgement.} This research is supported by the National Key R\&D Program of China (2024YFE0217700), National Natural Science Foundation of China (62472184) and the Fundamental Research Funds for the Central Universities.
% \end{sloppypar}
% \section{Conclusion} We propose PriOr-Flow, a novel and universal dual-branch panoramic optical flow estimation method. The proposed Dual-Cost Collaborative Lookup (DCCL) operator and Ortho-Driven Distortion Compensation (ODDC) module leverage the low-distortion prior of orthogonal views to enhance the network's performance in the polar regions. Our PriOr-RAFT achieves state-of-the-art performance on both the MPF Dataset and FlowScape, and our method also demonstrates strong performance in real-world scenarios.

% \section{Conclusion}
% \vspace{-5pt}
% \begin{sloppypar}
% We propose PriOr-Flow, a novel and universal dual-branch method for panoramic optical flow estimation. By introducing the Dual-Cost Collaborative Lookup (DCCL) operator and the Ortho-Driven Distortion Compensation (ODDC) module, PriOr-Flow leverages the low-distortion prior from the orthogonal view to significantly improve performance, especially in polar regions. Our PriOr-RAFT model achieves state-of-the-art performance on both the MPFDataset and FlowScape, with notable improvements in accuracy. Additionally, PriOr-Flow demonstrates strong generalization in real-world scenarios, validating its effectiveness beyond synthetic datasets.
% \end{sloppypar}

% \begin{sloppypar}
% \noindent\textbf{Acknowledgement.} This research is supported by the National Key R\&D Program of China (2024YFE0217700)，National Natural Science Foundation of China (62472184) and the Fundamental Research Funds for the Central Universities.
% \end{sloppypar}
{
    \small
    \bibliographystyle{ieeenat_fullname}
    \bibliography{main}
}

\end{CJK}
\end{document}